\documentclass[]{dataflow}

\usepackage[toc,page,header]{appendix}

\usepackage{minitoc}
\usepackage{multirow}     
\usepackage{multicol}     
\usepackage{booktabs}     
\usepackage{colortbl}     
\usepackage[table]{xcolor} 
\usepackage{makecell}     
\usepackage{graphicx}     
\usepackage{minted}
\usepackage{xcolor}
\usepackage{fontawesome5}
\usepackage{listings}
\usepackage{xcolor}

\definecolor{codegreen}{rgb}{0,0.6,0}
\definecolor{codegray}{rgb}{0.5,0.5,0.5}
\definecolor{codepurple}{rgb}{0.58,0,0.82}
\definecolor{backcolour}{rgb}{0.96,0.96,0.96}

\lstdefinestyle{mystyle}{
    backgroundcolor=\color{backcolour},   
    commentstyle=\color{codegreen}\itshape,
    keywordstyle=\color{magenta}\bfseries,
    numberstyle=\tiny\color{codegray},
    stringstyle=\color{codepurple},
    basicstyle=\ttfamily\footnotesize, 
    breakatwhitespace=false,         
    breaklines=true,                 
    captionpos=b,                    
    keepspaces=true,                 
    numbers=left,                    
    numbersep=5pt,                  
    showspaces=false,                
    showstringspaces=false,
    showtabs=false,                  
    tabsize=4,
    frame=single,                    
    rulecolor=\color{black!30}       
}

\title{OpenWorldLib: A Unified Codebase and Definition of Advanced World Models}

\author[]{DataFlow Team$^{1}$}
\author[\dagger]{Bohan Zeng$^{1,2}$}
\author[*]{Daili Hua$^{1}$}
\author[*]{Kaixin Zhu$^{1}$}
\author[*]{Yifan Dai$^{7,2}$}
\author[*]{Bozhou Li$^{1,2}$}
\author[*]{Yuran Wang$^{1,2}$}
\author[*]{Chengzhuo Tong$^{1,2}$}
\author[*]{Yifan Yang$^{1}$}
\author[*]{Mingkun Chang$^{8}$}
\author[*]{Jianbin Zhao$^{1}$}
\author[]{Zhou Liu$^{1}$}
\author[]{Hao Liang$^{1}$}
\author[]{Xiaochen Ma$^{4}$}
\author[]{Ruichuan An$^{1}$}
\author[]{Junbo Niu$^{1}$}
\author[]{Zimo Meng$^{1}$}
\author[]{Tianyi Bai$^{4}$}
\author[]{Meiyi Qiang$^{1}$}
\author[]{Huanyao Zhang$^{1}$}
\author[]{Zhiyou Xiao$^{1}$}
\author[]{Tianyu Guo$^{1}$}
\author[]{Qinhan Yu$^{1}$}
\author[]{Runhao Zhao$^{1}$}
\author[]{Zhengpin Li$^{1}$}
\author[]{Xinyi Huang$^{1}$}
\author[]{Yisheng Pan$^{1}$}
\author[]{Yiwen Tang$^{1}$}
\author[]{Juanxi Tian$^{10}$}
\author[]{Yang Shi$^{1,2}$}
\author[]{Yue Ding$^{2}$}
\author[]{Xinlong Chen$^{2}$}
\author[]{Hongcheng Gao$^{5}$}
\author[]{Minglei Shi$^{5}$}
\author[]{Jialong Wu$^{5}$}
\author[]{Zekun Wang$^{2}$}
\author[]{Yuanxing Zhang$^{2}$}
\author[]{Xintao Wang$^{2}$}
\author[]{Pengfei Wan$^{2}$}
\author[]{Yiren Song$^{6}$}
\author[]{Mike Zheng Shou$^{6}$}
\author[\ddagger]{Wentao Zhang$^{1,3,9}$}

\affiliation[]{$^{1}$Peking University}
\affiliation[]{$^{2}$Kling Team, Kuaishou Technology}
\affiliation[]{$^{3}$Zhongguancun Academy}
\affiliation[]{$^{4}$HKUST}
\affiliation[]{$^{5}$Tsinghua University}
\affiliation[]{$^{6}$National University of Singapore}
\affiliation[]{$^{7}$Shanghai Jiao Tong University}
\affiliation[]{$^{8}$Sun Yat-sen University}
\affiliation[]{$^{9}$Beijing Key Laboratory of Data Intelligence and Security (Peking University)}
\affiliation[]{$^{10}$Nanyang Technological University}

\abstract{
World models have garnered significant attention as a promising research direction in artificial intelligence, yet a clear and unified definition remains lacking. In this paper, we introduce OpenWorldLib, a comprehensive and standardized inference framework for Advanced World Models. Drawing on the evolution of world models, we propose a clear definition: a world model is a model or framework centered on perception, equipped with interaction and long-term memory capabilities, for understanding and predicting the complex world. We further systematically categorize the essential capabilities of world models. Based on this definition, OpenWorldLib integrates models across different tasks within a unified framework, enabling efficient reuse and collaborative inference. Finally, we present additional reflections and analyses on potential future directions for world model research.

}

\date{\today}

\def\emailicon{\raisebox{-1.5pt}{\includegraphics[height=1.05em]{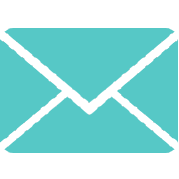}}}
\def\githubicon{\raisebox{-1.5pt}{\includegraphics[height=1.05em]{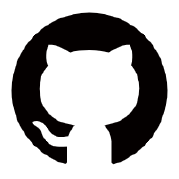}}}

\contribution[*]{Equal Contribution}
\contribution[\dagger]{Project Leader}
\contribution[\ddagger]{Corresponding author}

\checkdata[ \emailicon \hspace{0.3em} Correspondence ]{\email{wentao.zhang@pku.edu.cn}}

\newcommand{\sourcelink}{https://github.com/OpenDCAI/OpenWorldLib}

\checkdata[ \githubicon \hspace{0.3em} Source Code ]{ \url{\sourcelink} }

\checkdata[ \faFile \hspace{0.57em} Documentation ]{ \url{https://wcny4qa9krto.feishu.cn/wiki/XtPJwf5XQipP7RkeVv0ckyWlnNd} }

\begin{document}
\maketitle

\renewcommand{\thefootnote}{\fnsymbol{footnote}} 
\setcounter{footnote}{0}

\renewcommand{\thefootnote}{\arabic{footnote}}
\pagestyle{fancy}
\fancyhf{}
\fancyhead[L]{OpenDCAI Technical Report}
\fancyhead[R]{\thepage}

\newpage
\tableofcontents
\newpage

\section{Introduction}

\begin{figure}[htbp]
  \centering
   \centerline{\includegraphics[width=\linewidth]{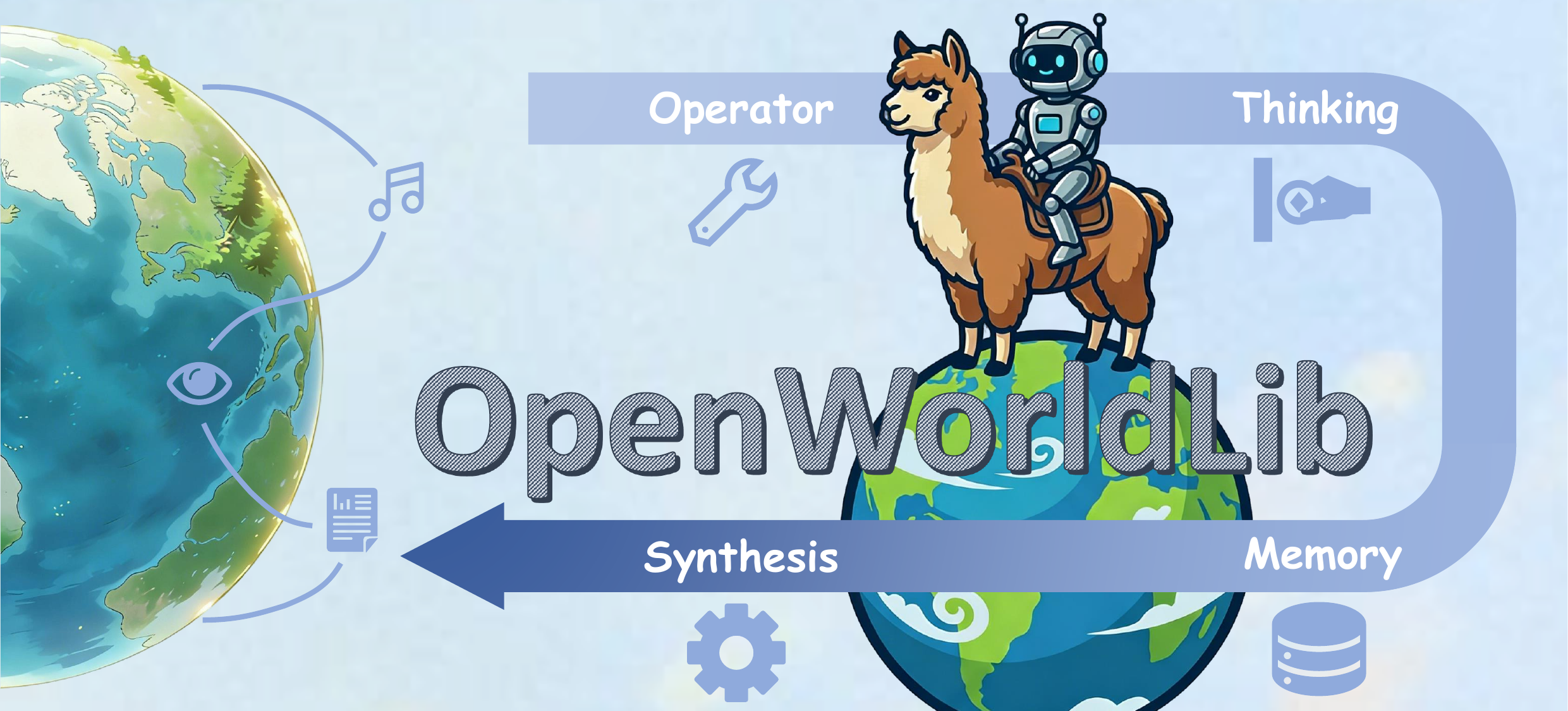}}
  \caption{Overview of our OpenWorldLib. Our OpenWorldLib establishes a unified framework for existing world model-related tasks, encompassing perception, understanding, memory, and generation of physical world inputs.}
  \label{fig:teaser}
\end{figure}

With the gradual advancement of LLMs and Agents~\cite{bai2023qwen, touvron2023llama, nie2025large, zhu2025llada, liu2024deepseek, shen2025let, wang2024qwen2, li2023blip2, shi2025mavors, liang2025multimodal, guo2025seed1, comanici2025gemini, guo2025deepseek, openai2025o3, openai2025gpt5, bai2025multi, guo2024deepseek, devlin2019bert, liang2025dataflow, liang2026dataflex, bai2024survey, an2026genius}, models are urgently required to transition from virtual-world usage to real-world applications. As a result, world models have begun to enter the spotlight, with researchers increasingly focusing on the ability of large models to function in the physical world, moving beyond virtual environments.

The concept of world models was initially introduced by \cite{ha2018world}, and later works such as \cite{hafner2023mastering, bardes2023v, assran2025v} began to next-frame-predict tasks like video generation and 3D generation as forms of world modeling. As the field has evolved, subsequent studies have explored the applications of world models, and numerous survey~\cite{zhu2024sora, hu2025simulating, ding2025understanding, wei2026trinity, kong20253d, lin2025exploring, yu2025survey, li2025comprehensive, long2025survey, fung2025embodied, liu2025aligning, yang2025survey, tu2025role, feng2025survey, baraldi2025safety, huang2025awesomeworldmodels, yue2025simulating} and position~\cite{lecun2022path, xing2025critiques, bai2025positional, wu2026visual, zeng2026research} papers have provided summaries and analysis. However, despite these efforts, the definition and scope of world models remain diverse, and a broadly accepted consensus has yet to be established.

To provide a standardized definition of world models, it is helpful to start with their core objective: the ability to continuously learn from and interact with the real world. Accordingly, we define a world model as \textbf{a model or framework centered on building internal representations from perception, equipped with action-conditioned simulation and long-term memory capabilities, for understanding and predicting the dynamics of a complex world}. As noted in~\cite{zeng2026research}, a world model is not tied to any specific task or particular architecture, but rather represents a level of capability that a model or framework should aim to achieve, specifically, the ability to perceive, interact with, and remember a complex world.

This article primarily defines which tasks fall within the capabilities a world model should possess, and which tasks are often mistakenly considered as what world models should achieve. Moreover, because world models require a diverse set of capabilities, a more systematic approach to invoking them is needed. To this end, we build OpenWorldLib, a unified world model inference framework that standardizes the invocation of tasks such as interactive video generation, 3D generation, multimodal reasoning, and vision-language-action (VLA) under a single framework.

The main contributions of this paper are as follows:
\begin{itemize}

\item \textbf{Contribution~1.}
We provide a standardized definition of world models, clarifying which tasks should be considered part of a world model’s capabilities.

\item \textbf{Contribution~2.}
We propose OpenWorldLib, a unified world model inference framework, to help structure and standardize research in this area.

\item \textbf{Contribution~3.}
We offer further analysis and discussion on the future development of world models.

\end{itemize}


\section{Background and Related Works}
\label{sec:relatedwork}

World models~\cite{ha2018world,hafner2019dream} are typically defined by three core conditional probability distributions:
\begin{equation}
\begin{aligned}
& \text{State transition model:} \quad p(s_{t+1} \mid s_t, a_t) \\
& \text{Observation model:} \quad p(o_t \mid s_t) \\
& \text{Reward model:} \quad r_t \sim p(r_t \mid s_t, a_t)
\end{aligned}
\label{eq:world_model_original}
\end{equation}

where \(s_t\) denotes the latent state, which intrinsically incorporates memory storage to manage long-horizon dependencies for complex tasks; \(a_t\) represents the action at time step $t$, drawn from an action space that has been broadened to encompass diverse operations and task-specific outputs such as generation and manipulation; \(o_t\) is the perceptual observation (e.g., vision, audio, or proprioception); and \(r_t\) is the reward obtained through interactions between actions and the environment.

Despite the wide use of these formulations, many tasks formally satisfy such conditional probability distributions without actually serving the core purpose of world models. These tasks are often conflated with or loosely labeled as world model research. Therefore, in this section, we draw on definitions proposed in prior work alongside the perspective advocated in this paper to clearly delineate which tasks fall within the scope of genuine world model research and which do not.

\subsection{World Model Related Tasks}
\paragraph{Interactivate Video Generation.}
Next-frame prediction is widely regarded the most recognized paradigm by world model researchers~\cite{ha2018world}, establishing interactive video generation as the main focus of research in this field. Early approaches primarily relied on regression-based models~\cite{ha2018world, hafner2023mastering, wu2023pre, ma2024harmonydream} to predict subsequent frames. More recently, the field has shifted towards leveraging diffusion models~\cite{ho2022imagen, huang2026vidworld} to achieve higher-quality interactive video generation, with unified multimodal approaches~\cite{wu2024ivideogpt, wurlvr, cui2025emu3, li2026semantic, wang2025scone} further advancing generation fidelity and controllability. With the accelerated inference speeds of diffusion models, game video generation~\cite{li2025hunyuan, tang2025hunyuan, wu2026infinite} and camera-controlled video generation~\cite{bai2025recammaster, luo2025camclonemaster} has emerged as a particularly prominent area of interest. Furthermore, the video prediction paradigm has been successfully integrated into Vision-Language-Action (VLA) models~\cite{black2024pi_0, intelligence2025pi_, spiritai2026spiritv15} and autonomous driving systems~\cite{zeng2025rethinking, hu2023gaia, russell2025gaia}. By incorporating next-frame prediction estimation, these models achieve significantly enhanced stability and robustness in their predictive capabilities. However, while interactive video generation remains a cornerstone of current world model research, it is important to note that next-frame prediction is not the sole implementation paradigm. Considering that the ultimate objective of a world model is to facilitate long-term interactions within complex environments, exploring alternative or complementary representational paradigms is equally crucial \cite{wu2026visual}.

\paragraph{Multimodal Reasoning.}
A critical capability of a world model lies in its profound understanding of the complex physical world; thus, multimodal reasoning serves as a key reflection of a world model's capabilities. Multimodal reasoning tasks closely associated with world models encompass not only spatial reasoning~\cite{ma2025spatialreasoner, li2025spatialladder, wu2025spatial, chen2024spatialvlm, sarch2025grounded, song2025robospatial, yang2025thinking, gholami2025spatial,deitke2025molmo} and omni reasoning~\cite{xu2025qwen2, xu2025qwen3, wiedemer2025video, huang2025mllms, chen2025think, zheng2025learning, sun2025mm, zhou2024mathscape, lin2025perceiveanythingrecognizeexplain, an2025unictokens, chen2026diadem, ding2026omnisift}, but also temporal reasoning~\cite{nhu2025time, niu2025ovo, cai2025lovr, liang2024evqascore, liang2024keyvideollm}, causal reasoning~\cite{kiciman2023causal,foss2025causalvqa}. Recently, beyond traditional explicit reasoning methods, utilizing latent reasoning~\cite{assran2025v,Monet} to analyze complex dynamics in the real world has emerged as a prominent research hotspot. By shifting away from the traditional text-centric pre-training paradigms of Large Language Models (LLMs), latent reasoning mechanisms enable models to more efficiently ingest and process the complex, high-dimensional, and continuous information inherent in the real world.

\paragraph{Vision-Language-Action.}
The ultimate goal of a world model is to enable agents to interact with the physical world, and embodied devices serve as the primary representatives for interacting with complex environments. Consequently, Vision-Language-Action (VLA) has become a crucial capability that world models must support. In the domain of robotic arm manipulation, recent research primarily follows two trajectories: utilizing Multimodal Large Language Models (MLLMs) to directly predict actions~\cite{black2024pi_0, intelligence2025pi_, spiritai2026spiritv15, team2025gigabrain, zhai2025igniting, lin2025evo0visionlanguageactionmodelimplicit, assran2025v, cen2025worldvla, goswami2025osvi}, or combining action prediction with video generation~\cite{ali2025world, team2025gigaworld, li2026lingbotva, chi2025wow, wang2026hand2world, mao2025robot} to facilitate action planning through future frame prediction. Furthermore, this VLA paradigm is being broadly applied to more complex embodied scenarios, including mobile robots with highly complex and difficult-to-control dynamics~\cite{huang2025mobilevla, li2025momagen}, and autonomous driving systems operating in vastly broader environments~\cite{Zhang_2025_CVPR, hu2023gaia, russell2025gaia, zhu2025worldsplat, li2025omninwm, wang2024drivedreamer, zhao2025drivedreamer, li2024drivingdiffusion, zhang2024bevworld}, thereby advancing the closed-loop interactive capabilities of models in the real world.

\subsection{The Role of 3D and Simulators in World Models}
Beyond tasks that rely on directly observable perception, a key part of world models involves processing virtual environments. To ensure that physical space remains consistent during long-term interactions, researchers often use simulators to let models learn in a structured way. While interactive video generation creates a visual guess of the future, 3D representations provide a verifiable environment where physical rules can be strictly followed~\cite{li20252, liu2025worldmirror, zhao2025deepmesh, li2026worldgrow, tochilkin2024triposr, wang2025evoworld, lin2025partcrafter, zeng2024ipdreamer, yang2024semantic, zeng2024trans4d}.

In this context, 3D generation and reconstruction are essential for keeping a stable world state. Recent works like VGGT~\cite{wang2025vggt}, InfiniteVGGT~\cite{yuan2026infinitevggt}, and OmniVGGT~\cite{peng2025omnivggt} use visual geometry grounded transformers to link image inputs with real geometric structures. To handle continuous data from the real world, some models now maintain a persistent 3D state~\cite{wang2025continuous} or utilize hybrid memory for long-context reconstruction~\cite{zhang2026loger}, making sure the environment stays the same even when the agent moves. Additionally, new methods in metric 3D reconstruction~\cite{keetha2025mapanything}, depth estimation~\cite{lin2025depth}, and large view synthesis~\cite{jin2024lvsm} allow world models to recover exact physical spaces from any camera angle. By learning permutation-equivariant visual geometry~\cite{wang2025pi}, these models can work better across different types of physical settings.

Furthermore, simulators act as a "sandbox" for world models, helping them move from abstract thinking to real physical action. For these simulators to work in real-time, fast scene generation is necessary. For example, FlashWorld~\cite{li2025flashworld} and the Hunyuan series~\cite{team2025hunyuanworld, yang2024hunyuan3d, zhao2025hunyuan3d, hunyuan3d2025hunyuan3d} can create high-quality 3D scenes or assets in a very short time, giving the world model an immediate place to test its ideas. Recent investigations also explore the potential of reinforcement learning in these 3D generation processes~\cite{tang2025we}. By using these explicit 3D representations and simulation tools, world models can go beyond just predicting pixels and truly understand the physical rules of the real world.

\subsection{Methods Not Considered World Models}
Besides world model-related tasks, certain applications do not truly reflect a world model's capabilities, yet they frequently appear in similar discussions. Based on the formulation and our specific definition of world models, this section clarifies which tasks fall outside this category.

A prominent example of this misconception is text-to-video generation. When Sora was released, many label it a "world simulator." However, \cite{zhu2024sora} argues that Sora does not constitute a complete world simulator. While next-frame prediction frequently associates with world models, our definition emphasizes that the key lies not in the output format, but in whether the model utilizes multimodal inputs to analyze and recognize the environment. Next-frame prediction serves merely as one format. What truly matters is whether the model accurately understands complex physical rules and interacts with the world. Text-to-video generation lacks this complex perceptual input. Even though generating videos demonstrates some understanding of physics, it remains outside the core tasks of world models.

Similarly, some tasks, such as the code generation or web search~\cite{copet2025cwm, feng2025web}, borrow the long-term interaction structure of world models for other domains. However, these tasks typically lack multimodal inputs and do not involve understanding the physical world. While applying this structure to new areas presents interesting opportunities, these tasks do not qualify as true world models.

Even applications that actually involve multimodal ~\cite{dong2026mineru} and long-term interaction, such as avatar video generation~\cite{huang2025live, zeng2023face, guo2024liveportrait}, do not necessarily fit the definition. These tasks primarily focus on entertainment. Because they have little to do with exploring or understanding the complex physical world, they do not represent a primary focus for world models.


\begin{figure}[htbp]
  \centering
  \centerline{\includegraphics[width=0.9\linewidth]{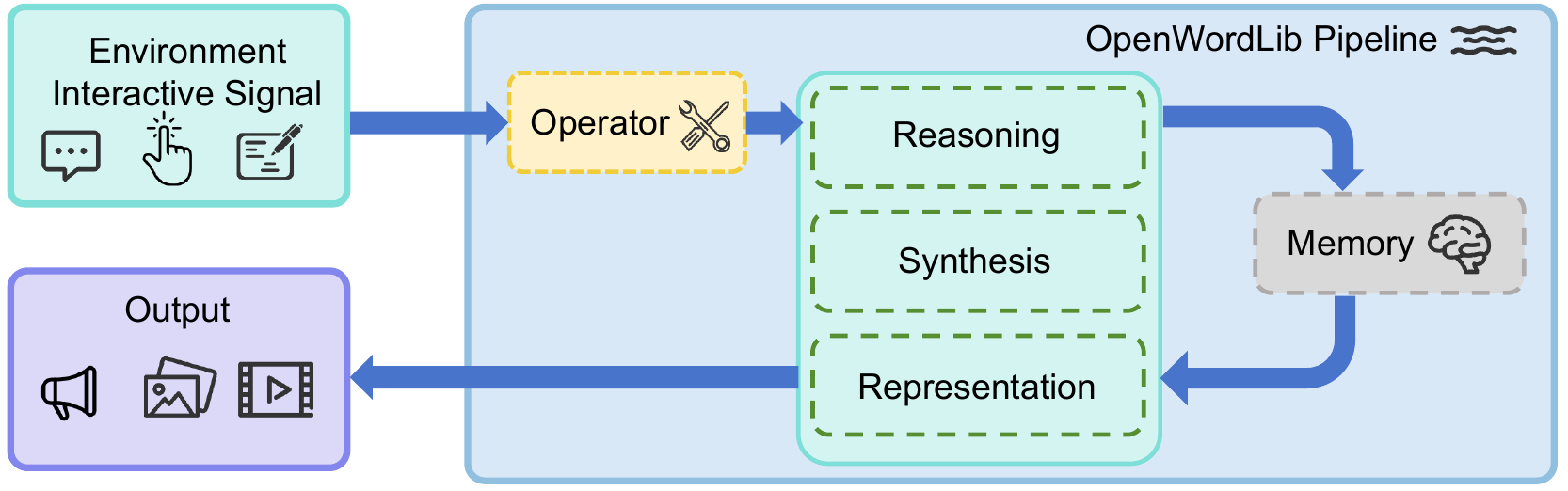}}
  \caption{Illustration of our OpenWorldLib framework.}
  \label{fig:framework_structure}
\end{figure}


  
  
  

\begin{figure}[htbp]
  \centering
  \centerline{\includegraphics[width=1.0\linewidth]{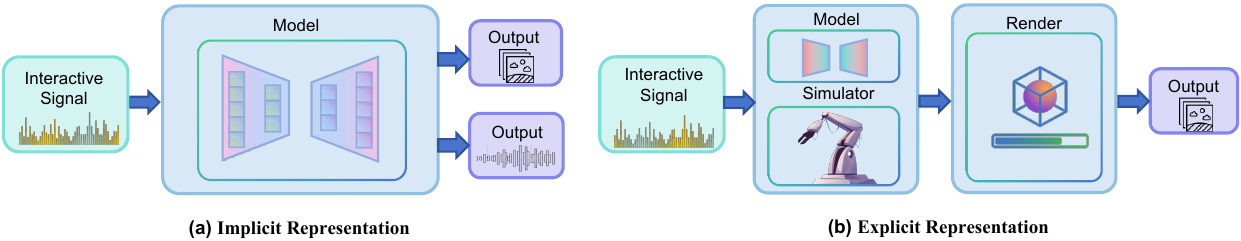}}
  \caption{Demonstration of world model Implicit representation and explicit representation.}
  \label{fig:framework_types}
\end{figure}

\section{OpenWorldLib Framework Design}
Based on Section~\ref{sec:relatedwork}, a world model requires the following capabilities: receiving inputs from the complex physical world, understanding the physical world, maintaining long-term memory during interactions, and supporting multimodal outputs. Although \cite{zeng2026research} proposed a design for a unified world model framework, it lacks a concrete engineering implementation and even a unified standard. This section details the specific design of our OpenWorldLib framework, as shown in Fig.~\ref{fig:framework_structure}.

\subsection{Operator}
In the OpenWorldLib framework, the \textit{Operator} module serves as the crucial bridge between raw user inputs (or environmental signals) and the core execution modules (Synthesis, Reasoning, and Representation). Because a world model must handle complex, multimodal inputs from the physical world—such as text prompts, images, continuous control actions, and audio signals—the Operator is designed to standardize these diverse data streams. 

Specifically, when the \texttt{Pipeline} is called, it routes the raw input through the Operator's \texttt{process()} method. The Operator is responsible for two primary functions:
\begin{itemize}
    \item \textbf{Validation:} Ensuring that the input data formats, shapes, and types meet the requirements of the downstream models.
    \item \textbf{Preprocessing:} Transforming raw signals into standardized tensor representations or structured formats (e.g., resizing images, tokenizing text, or normalizing action spaces).
\end{itemize}

To facilitate the integration of new world model methods, we define a unified \texttt{Operator} template. All task-specific operators inherit from this base class, ensuring a consistent API across the entire codebase. The definition of \texttt{Operator} is shown in Listing~\ref{lst:base_operator}.

\begin{lstlisting}[language=Python, caption={The template definition of BaseOperator in OpenWorldLib.}, label={lst:base_operator}]
from typing import Any, Dict, Union


class BaseOperator(object):
    def __init__(self):
        self.current_interaction = []
        self.interaction_template = []


    def get_interaction(self, interaction_list):
        for act in interaction_list:
            self.check_interaction(act)
        self.current_interaction.append(interaction_list)

    def check_interaction(self, interaction):
        if interaction not in self.interaction_template:
            raise ValueError(f"{interaction} not in template")
        return True

    def process_interaction(self):
        raise NotImplementedError("Subclasses must implement process_interaction().")

    def process_perception(self):
        raise NotImplementedError("Subclasses must implement process_perception().")

\end{lstlisting}

\subsection{Synthesis Module}


As shown in the implicit representation part of Fig.~\ref{fig:framework_types}, a core capability of world models is using internal learned dynamics to generate visual, auditory, and other sensory outcomes as environmental feedback. We define this implicit generative process as the model's \textit{implicit representation}.
In the OpenWorldLib framework, the \textit{Synthesis} module serves as the generative bridge between standardized conditioning from upstream pipelines and the multimodal outputs(visual, auditory, and embodied) that users, simulators, or robotic stacks actually consume. Because a world model must realize predictions not only as internal states but as observable media and executable commands, Synthesis hosts heterogeneous generative backends while preserving a coherent integration pattern across modalities.

Specifically, when the \texttt{Pipeline} runs a generation path, it passes operator-aligned inputs to the appropriate synthesis backend, which performs inference under modality-specific controls and returns structured artifacts together with concise metadata for export, evaluation, or memory. The following subsections unpack this module along its visual, audio, and other physical-signal synthesis branches.

\begin{lstlisting}[language=Python, caption={The template definition of BaseSynthesis for visual (and other generative) backends in OpenWorldLib.}, label={lst:base_synthesis}]
import torch

class BaseSynthesis(object):
    def __init__(self):
        """
        Initialize the synthesis model, including inference components,
        auxiliary modules, and processing utilities.
        """
        pass

    @classmethod
    def from_pretrained(cls, pretrained_model_path, args, device=None, **kwargs):
        """
        Load pretrained model weights and return an inference-ready synthesis instance.
        """
        pass

    def api_init(self, api_key, endpoint):
        """
        Initialize API key and endpoint for cloud-based synthesis services.
        """
        pass

    @torch.no_grad()
    def predict(self):
        """
        Run model inference to produce generated outputs such as images,
        video frames, or other synthesized media.
        """
        pass
\end{lstlisting}

\subsubsection{Visual Synthesis}
The visual synthesis layer covers image and video oriented generation in OpenWorldLib: it turns structured conditioning, such as text prompts, reference images, or scene-level specifications, into raster outputs (frame tensors, decoded clips, or API-returned assets) together with metadata for export, evaluation, or optional memory hooks. 
In this way, the framework can furnish visible predictions of how scenes evolve over time, which is essential for interactive simulation, qualitative inspection, and comparing alternative futures or camera paths at a glance.
These outputs also anchor multimodal storytelling and dataset-style recording when visual evidence must accompany text or control signals.

In practice, the visual synthesis layer is organized around the following responsibilities:
\begin{itemize}
    \item \textbf{Generative stack composition:} Combining text encoders, latent decoders, and diffusion- or flow-based cores with schedulers or solvers appropriate to each task, and exposing knobs for spatial resolution, temporal extent (frame budget), and guidance-style parameters.
    \item \textbf{Integration surfaces:} Supporting checkpoint-driven pipelines (unified construction from pretrained resources and no-gradient inference) alongside hosted-service wrappers that authenticate via endpoints and credentials, so that local and remote generators share the same conceptual call pattern.
\end{itemize}

\subsubsection{Audio Synthesis}
The audio synthesis layer focuses on continuous waveform generation under structured conditioning, commonly text, optional video-derived features, and timing or batch metadata, and returns waveforms with sampling rates and compact result records for downstream saving or metrics. 
Its role is to supply the auditory side of multimodal outputs so that scenarios are not limited to silent video or text-only feedback, which matters for perception-rich environments and for judging alignment between sound and visuals~\cite{guo2025brace}.

Concretely, the audio synthesis layer fulfills the following roles:
\begin{itemize}
    \item \textbf{Resource assembly:} Instantiating the neural audio generator and any auxiliary modules (e.g., feature encoders) from pretrained sources through a single factory-style entry point, with explicit device and reproducibility-related settings.
    \item \textbf{Conditional waveform synthesis:} Mapping operator-prepared tensors and prompts to audio outputs via a unified inference entry point, with user-facing controls such as duration, random seeds, guidance strength, and sampling-step budgets.
\end{itemize}

\subsubsection{Other Signal Synthesis}
Beyond visual and audio modalities, comprehensive interaction with the environment requires world models to generate a diverse spectrum of physical signals. Among these, action control proves especially critical, as it constitutes the fundamental mechanism through which embodied agents actively manipulate the physical world. OpenWorldLib therefore places a primary emphasis on Vision-Language-Action (VLA) signal generation within this module.
This synthesis layer is tailored for embodied tasks and fulfills the following functions:
\begin{itemize}
\item \textbf{Policy Initialization and Space Alignment:} Loading specialized physical policies (e.g., VLA foundation models) from pretrained weights. Crucially, it maps diverse action representations—from discrete language-like tokens to continuous kinematic states—into unified interfaces compatible with target simulators or robotic hardware.
\item \textbf{Context-Conditioned Action Synthesis:} Translating rich, multimodal contexts (such as real-time visual streams, textual goals, and proprioceptive histories) into grounded physical commands. The module yields executable action sequences and essential control metadata to drive closed-loop environmental interactions.
\end{itemize}

\subsection{Reasoning Module}

From the implicit representation part of Fig.~\ref{fig:framework_types}, a world model must transcend mere perception to understand the physical world: inferring spatial relationships, integrating multimodal contexts, and generating grounded semantic interpretations prior to any downstream generation or action takes place. To address this necessity, OpenWorldLib introduces a dedicated Reasoning module, designed to equip the world model with structured understanding capabilities essential for complex physical inference.
Concretely, the Reasoning module is organized into three sub-categories that reflect the distinct perceptual channels a world model must handle:
\begin{itemize}
    \item \textbf{General Reasoning:} Multimodal large language models (MLLMs) capable of processing text, images, audio, and video in a unified manner.
    \item \textbf{Spatial Reasoning:} Models specialized in 3D spatial understanding and object localization from visual observations.
    \item \textbf{Audio Reasoning:} Models that interpret and reason over auditory signals.
\end{itemize}

To facilitate the integration of new reasoning-oriented world model methods, we define a unified \texttt{BaseReasoning} template. All task-specific reasoning classes inherit from this base class, ensuring a consistent API across the entire codebase. The definition of \texttt{BaseReasoning} is shown in Listing~\ref{lst:base_reasoning}.

\begin{lstlisting}[language=Python, caption={The template definition of BaseReasoning in OpenWorldLib.}, label={lst:base_reasoning}]
import torch

class BaseReasoning(object):
    def __init__(self):
        """
        Initialize the reasoning model, equipping it with structured 
        understanding capabilities for general, spatial, or audio inference.
        """
        pass

    @classmethod
    def from_pretrained(cls, pretrained_model_path, device=None, **kwargs):
        """
        Load pretrained model weights and return an inference-ready reasoning instance.
        """
        pass

    def api_init(self, api_key, endpoint):
        """
        Initialize API key and endpoint for cloud-based reasoning services.
        """
        pass

    @torch.no_grad()
    def inference(self):
        """
        Run model inference to integrate multimodal contexts, infer spatial 
        relationships, and generate grounded semantic interpretations.
        """
        pass
\end{lstlisting}

\subsection{Representation Module}

Apart from models that use internal capabilities to understand the world, some methods aim to build human-defined simulators, such as 3D meshes. These simulators provide a testable environment for the world model framework. Since these structured representations are different from the perception data that can be directly collected from the world, we design the Representation module separately from the Synthesis module to handle these explicit representations (e.g., 3D structures).

Specifically, the Representation module is designed to bridge the gap between raw perception and structured simulation. Its main functions include:
\begin{itemize}
    \item \textbf{3D Reconstruction:} It transforms input data into explicit 3D outputs, providing structured information such as point clouds, depth maps, and camera poses.
    \item \textbf{Simulation Support:} It creates a manual environment where the world model can test its reasoning and validate if its predicted actions are correct in a coordinate system.
    \item \textbf{Service Integration:} It supports both local inference and cloud-based APIs to help export these explicit representations to external physics engines.
\end{itemize}

To standardize how these models are used, we provide a unified \texttt{BaseRepresentation} template. All task-specific representation classes inherit from this base class to ensure a consistent API. The definition of \texttt{BaseRepresentation} is shown in Listing~\ref{lst:base_representation}.

\begin{lstlisting}[language=Python, caption={The template definition of BaseRepresentation in OpenWorldLib.}, label={lst:base_representation}]
class BaseRepresentation(object):
    def __init__(self):
        """
        Initialize the representation model, including architecture
        setup, device placement, and dtype configurations.
        """
        pass

    @classmethod
    def from_pretrained(cls, pretrained_model_path, device=None, **kwargs):
        """
        Load pretrained model weights and return an inference-ready instance.
        """
        pass

    def api_init(self, api_key, endpoint):
        """
        Initialize API key and endpoint for cloud-based model services.
        """
        pass

    @torch.no_grad()
    def get_representation(self, data):
        """
        Run model inference to produce 3D outputs such as points, 
        depth maps, camera poses, or masks.
        """
        pass
\end{lstlisting}

\subsection{Memory Module}
Long-term contextual memory is essential for interactive world models to maintain historical observations, reasoning chains, and interaction states. OpenWorldLib designs a unified \textit {Memory} module to manage multimodal interaction history.

The Memory module serves as the persistent state center of the framework. It records structured information from perception, reasoning, generation, and action, and provides efficient context retrieval for multi-turn interactive tasks. Specifically, the Memory module fulfills the following functions:
\begin{itemize}
\item \textbf{Historical Storage:} Storing text, visual features, action trajectories, and scene states across interactions.

\item \textbf{Context Retrieval:} Selecting relevant history to support consistent reasoning and generation.

\item \textbf{State Update:} Recording new interaction results after each pipeline execution.

\item \textbf{Session Management:} Supporting independent memory for different tasks and sessions.
\end{itemize}

To unify memory management, we define a unified \texttt{BaseMemory} template. All task-specific memory classes inherit from this base class. The definition of \texttt{BaseMemory} is shown in Listing~\ref{lst:base_memory}.

\begin{lstlisting}[language=Python, caption={The template definition of BaseMemory in OpenWorldLib.}, label={lst:base_memory}]
class BaseMemory(object):
    """
    Generic Multimodal Memory System Template
    """
    def __init__(self, capacity=None, **kwargs):
        """
        Initialize memory storage structure
        """
        self.storage = []

    def record(self, data, metadata=None, **kwargs):
        """
        Store current interaction data and metadata into memory
        """
        pass

    def select(self, context_query, **kwargs):
        """
        Retrieve relevant memory based on current context
        """
        pass

    def compress(self, memory_items, **kwargs):
        """
        Compress and refine memory to reduce redundancy
        """
        pass

    def manage(self, **kwargs):
        """
        Manage memory lifecycle: update, merge, or clean stale data
        """
        pass
\end{lstlisting}

\subsection{Pipeline}
To integrate the Operator, Reasoning, Synthesis, Representation, and Memory modules into a cohesive and usable system, OpenWorldLib provides a unified \textit {Pipeline} module as the top-level scheduling and execution entry. The Pipeline encapsulates model initialization, data flow, module invocation, memory interaction, and result post-processing, enabling end-to-end world model inference with a simple, consistent API.

The Pipeline follows a standard forward-execution workflow: it receives raw user or environmental input, routes it to the Operator for validation and preprocessing, queries the Memory module for historical context, coordinates the Reasoning, Synthesis, and Representation modules for core computation, and finally returns structured outputs while updating the memory. This design fully decouples module implementations while ensuring efficient and reliable data transmission.
The core responsibilities of the Pipeline include:
\begin{itemize}
    \item \textbf{Unified Model Initialization:} Loading pretrained weights, configuring devices, and instantiating all submodules through a single \texttt{from\_pretrained()} interface.
    
    \item \textbf{End-to-End Inference:} Implementing one-click forward inference via the \texttt{\_\_call\_\_()} method for single-turn world model tasks.
    
    \item \textbf{Multi-turn Interactive Execution:} Supporting stateful, continuous interaction through the \texttt{stream()} method with automatic memory reading and writing.
    
    \item \textbf{Modular Orchestration:} Dynamically invoking Reasoning, Synthesis, or Representation according to task type without modifying internal module logic.
    
    \item \textbf{Result Structuring:} Organizing outputs into standardized formats for visualization, evaluation, logging, or downstream control systems.
\end{itemize}

To maintain framework-wide consistency, all task-specific pipelines inherit from a unified \texttt{BasePipeline} template. Its definition is shown in Listing~\ref{lst:base_pipeline}.
\begin{lstlisting}[language=Python, caption={The template definition of Pipeline in OpenWorldLib.}, label={lst:base_pipeline}]
import torch
from typing import Generator, List

class BasePipeline:
    def __init__(self):
        """
        Initialize core submodules (Operator, Memory, Reasoning, etc.)
        """
        pass

    @classmethod
    def from_pretrained(cls):
        """
        Load pretrained weights and return a pipeline instance
        """
        return cls()
    
    def process(self, *args, **kwds):
        """
        Input validation and preprocessing (via Operator module)
        """
        pass
    
    def __call__(self, *args, **kwds):
        """
        Single-turn forward inference for world model tasks
        """
        pass

    def stream(self, *args, **kwds) -> Generator[torch.Tensor, List[str], None]:
        """
        Multi-turn continuous interaction with persistent memory
        """
        pass
\end{lstlisting}

\section{Discussion}
OpenWorldLib is designed to provide a clearer and more standardized definition and framework for world models. Its goal is to promote the development of world models so that AI can better assist humans in complex environments. In this section, we discuss the future development directions of world models.

Many current world model architectures focus on next-frame prediction. This approach aligns with how humans process high-density sensory inputs, as humans are essentially "pre-trained" in the physical world, whereas large models are pre-trained on massive internet text corpora~\cite{liang2026towards, liang2026dataprep}. However, based on existing architectures, VLMs might offer a practical solution. For example, Bagel~\cite{zhao2025unified} successfully achieves both multimodal reasoning and multimodal generation using the Qwen architecture. This demonstrates that Large Language Models (LLMs) pre-trained on internet data can possess all the capabilities required for a world model, showing their potential to serve as the foundational base. Therefore, before focusing entirely on the specific structural design of world models, we should first consider how to implement all their necessary functions to enable true and effective interaction with the complex world. Moreover, as LLMs serve as the foundational backbone for world models, data-centric methodologies—including multimodal data synthesis~\cite{liu2024synthvlm, liang2024synth}, domain-specific data augmentation~\cite{cai2025text2sql, zheng2024pas}, dynamic training~\cite{liang2026dataflex}, and training data quality evaluation~\cite{liang2025mathclean}—will play an increasingly important role in strengthening the foundational models that underpin world model capabilities.

During real-world interactions, next-frame prediction retains more information compared to next-token prediction, but its efficiency needs significant improvement. To boost this efficiency, improvements must start at the hardware level. Current computer byte organization naturally favors next-token prediction. Even when models attempt next-frame prediction, the data is still processed as tokens during actual computation. To achieve the ideal world model, we need hardware iterations, changes to the foundational model structure (token-based Transformers may need to evolve), and the comprehensive realization of complex physical world interaction tasks.

\begin{figure}[htbp]
  \centering
  \centerline{\includegraphics[width=\linewidth]{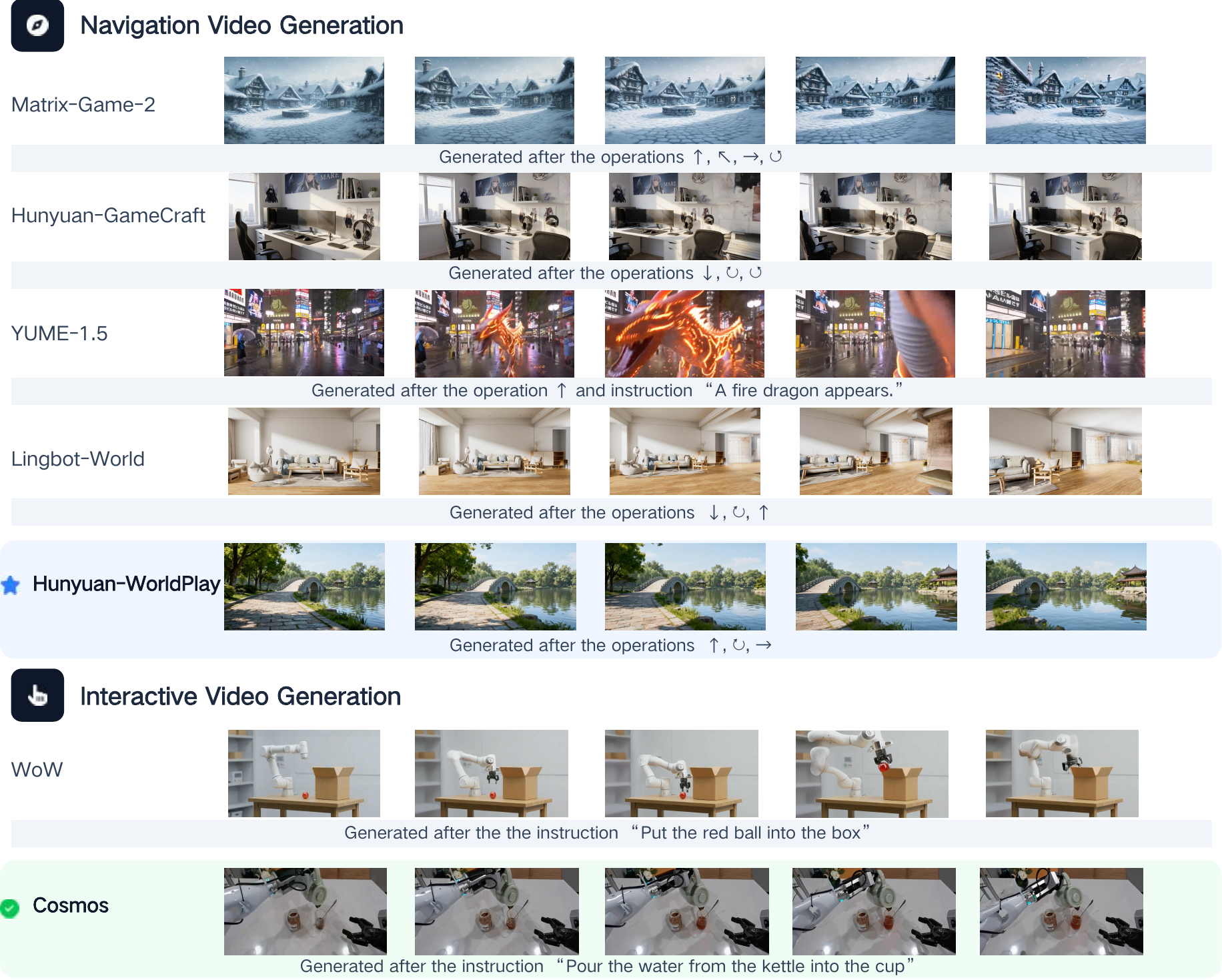}}
  \caption{Demonstration of interactive video generation results.}
  \label{fig:exp_video_gen}
\end{figure}

\begin{figure}[htbp]
  \centering
  \centerline{\includegraphics[width=\linewidth]{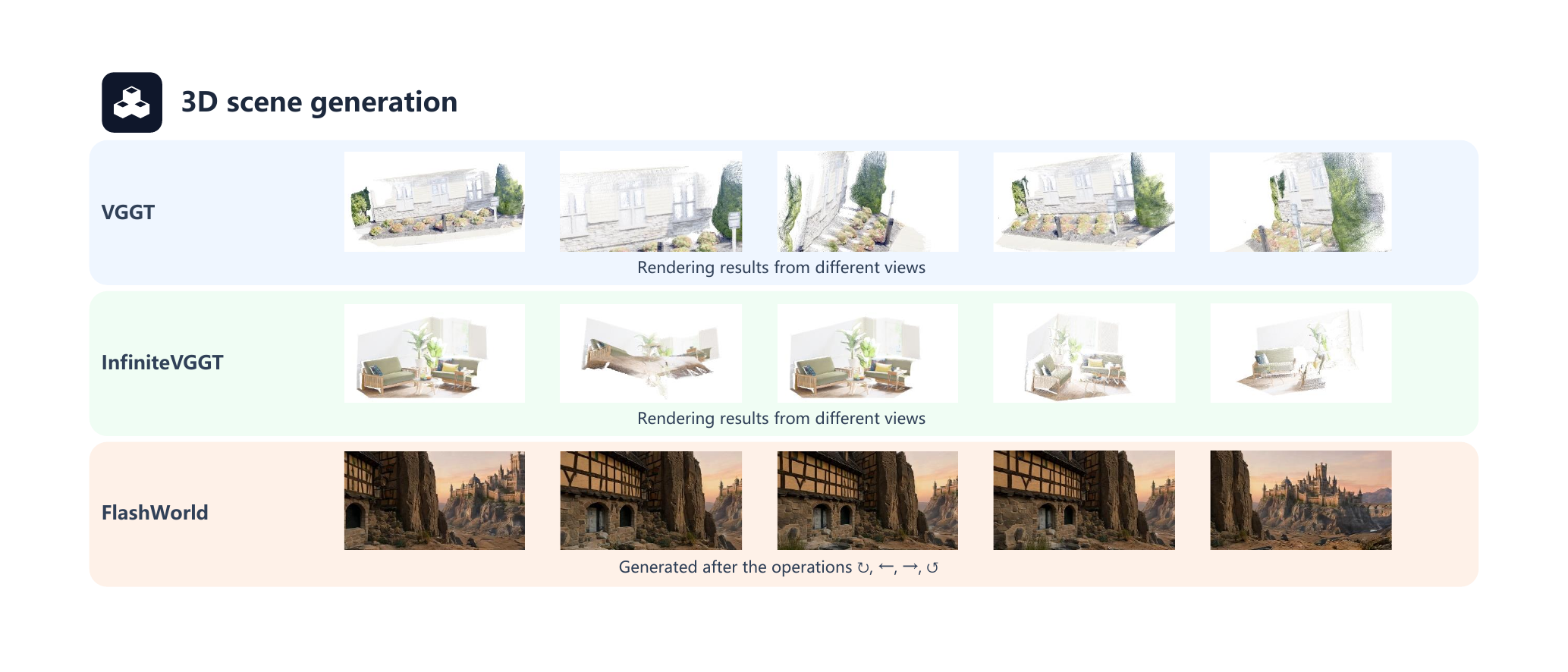}}
  \caption{Demonstration of 3D scene generation results.}
  \label{fig:3d_gen}
  \vspace{-3mm}
\end{figure}

\section{Evaluation}
Following the introduction of the OpenWorldLib framework design, this section presents the testing process and evaluation results of OpenWorldLib.

\subsection{Experimental Setting}
We primarily conduct our experiments using NVIDIA A800 (80GB VRAM) and H200 (141GB VRAM) GPUs. In the future, we plan to evaluate our framework on a wider range of hardware devices.

\subsection{Experimental Results}

\subsubsection{Interactive Video Generation.}
For video generation, our evaluation primarily encompasses tasks such as navigation video generation and interactive video generation. The significance of video generation for a world model lies in assessing its understanding and memory of the complex world, while simultaneously assisting other sequential reasoning tasks in making accurate predictions. When executing video generation tasks, the world model must produce videos that conform to the precise visual evolution required by the specific task.Specifically, the inputs for these tasks consist of visual conditions (e.g., single images or image sequences) paired with diverse interaction signals, which include textual instructions, directional movement controls (forward, backward, left, right), and camera rotation commands.

As shown in Fig.~\ref{fig:exp_video_gen}, we evaluate and analyze the generation performance of various methods. In the context of navigation video generation, early approaches like Matrix-Game-2~\cite{zhang2025matrix, he2025matrix} offer fast generation speeds but suffer from noticeable color shifting during long-horizon generation. In contrast, recent models such as Lingbot-World~\cite{team2026lingbotworld}, Hunyuan-GameCraft~\cite{li2025hunyuan, tang2025hunyuan}, and YUME-1.5~\cite{mao2025yume, mao2025yume1p5} successfully support high-quality navigation video generation, with Hunyuan-WorldPlay~\cite{sun2025worldplay} achieving the best overall visual performance. Regarding interactive video generation, although Wan-IT2V~\cite{wan2025wan} can execute basic interactive generation, it struggles with maintaining physical consistency. Furthermore, for generating complex interactive operations, while WoW~\cite{chi2025wow} supports a diverse range of functionalities, its generation quality and physical realism are significantly inferior to those of Cosmos~\cite{agarwal2025cosmos}.

\subsubsection{Multimodal Reasoning.}
In OpenWorldLib, the \textit{Reasoning} module groups high-level cognitive tasks that require a world model to interpret multimodal evidence and produce explicit, verifiable conclusions. It covers spatial reasoning \cite{ma2025spatialreasoner, li2025spatialladder} (e.g., answering geometry- and layout-centric queries, resolving object relations, and performing step-by-step spatial deductions from visual inputs) as well as omni/general reasoning \cite{xu2025qwen3} that operates over mixed modalities (text, images, audio, and videos) to support broad instruction following and multimodal understanding. The importance of this module for a world model lies in making internal perception and memory actionable: it turns observations into grounded decisions, explanations, and plans that can guide downstream generation or control. When executing reasoning tasks, inputs typically consist of an instruction or question paired with optional perceptual signals such as images, video clips, 
or audio segments, encoded into model-ready representations; the outputs are primarily decoded natural-language responses, and for certain omni reasoning settings may additionally include generated audio alongside the text.

\begin{figure}[htbp]
  \centering
  \centerline{\includegraphics[width=0.9\linewidth]{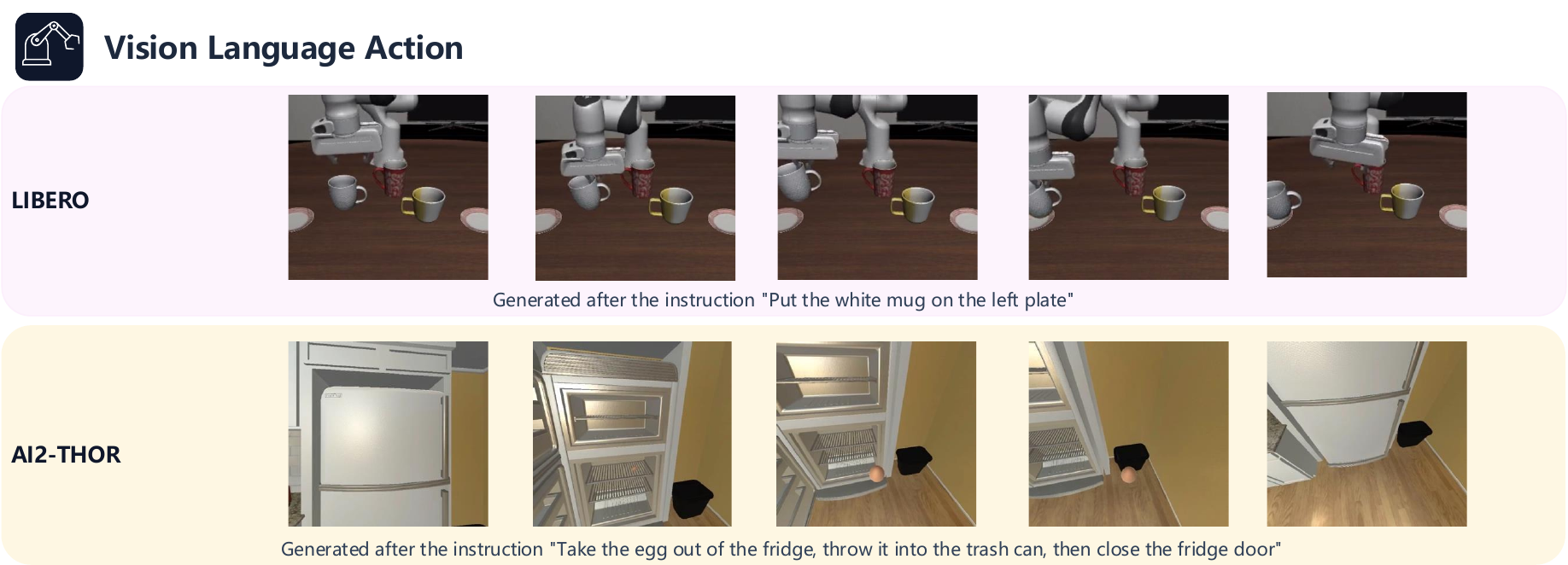}}
  \caption{Demonstration of simulator generation results.}
  \label{fig:simulator}
  \vspace{-3mm}
\end{figure}

\subsubsection{3D Generation.}
The 3D generation pipeline in OpenWorldLib supports 3D scene reconstruction, enabling robust representations of complex real-world environments. The perceptual inputs for this pipeline typically consist of single images or image sequences, while the interaction signals involve movement controls or camera viewpoint adjustments (e.g., polar, azimuth, and yaw angles). As shown in Fig.~\ref{fig:3d_gen}, although VGGT~\cite{wang2025vggt} and InfiniteVGGT~\cite{yuan2026infinitevggt} can generate 3D scenes from different views, they still have clear limitations. For instance, when the camera moves significantly, these models often struggle with geometric inconsistency and show texture blurring in complex areas, which affects the overall realism. While faster methods like FlashWorld~\cite{li2025flashworld} speed up the process, balancing steady shapes with sharp details remains a major challenge. Nevertheless, as a crucial technique for realistic physical simulation, 3D generation remains fundamentally important for the development of world models.

\subsubsection{Vision-Language-Action Generation.}

Similar to 3D generation, simulation environments constitute an indispensable component of world model evaluation, serving as controllable testbeds for both embodied video synthesis and action generation. To this end, OpenWorldLib incorporates two complementary simulation-based paradigms: AI2-THOR~\cite{kolve2017ai2} for embodied video generation, enabling photorealistic scene rendering and dynamic agent-environment interaction; and LIBERO~\cite{liu2023libero} for Vision-Language-Action (VLA) evaluation, providing reproducible and physically grounded manipulation environments. Together, these paradigms rigorously assess the world model's capacity to couple semantic understanding with physical dynamics and fine-grained action planning across diverse interactive scenarios.

As shown in Fig.~\ref{fig:simulator}, we present representative evaluation cases from both LIBERO and AI2-THOR simulation environments, showcasing diverse manipulation tasks and embodied interaction scenarios. Furthermore, our framework supports the evaluation of a comprehensive suite of VLA methods. Prominent examples include $\pi_0$~\cite{black2024pi_0} and $\pi_{0.5}$~\cite{intelligence2025pi_}, which leverage the PaliGemma vision-language backbone augmented with mixture-of-experts (MoE) action heads to achieve robust multi-task generalization. We also incorporate LingBot-VA~\cite{li2026lingbotva}, which approaches the task from a generative perspective by employing a video diffusion architecture to jointly model visual future predictions and continuous action synthesis.



\section{Conclusion}
In conclusion, OpenWorldLib presents a standardized workflow and evaluation pipeline for world models. By providing unified interfaces for core tasks such as interactive video generation and 3D scene reconstruction, the framework standardizes the integration of multimodal perceptual inputs and diverse interaction controls. Ultimately, we hope OpenWorldLib can serve as a practical reference for the research community, facilitating future explorations and fair comparisons in world model research.

\clearpage

\bibliographystyle{plainnat}
\bibliography{main}

\clearpage

\beginappendix

\section{Author Contributions}

\newcommand{\ProjectLeader}{\textcolor{red!70!black}{\textit{Project Leader}}}
\newcommand{\ProjectFounder}{\textcolor{blue!70!black}{\textit{Project Founder}}}
\newcommand{\CoreContributor}{\textcolor{green!50!black}{\textit{Core Contributor}}}
\newcommand{\Contributor}{\textcolor{green!80!black}{\textit{Contributor}}}
\newcommand{\ProjectSupervisor}{\textcolor{purple!70!black}{\textit{Project Supervisor}}}
\newcommand{\CorrespondingAuthor}{\textcolor{orange!80!black}{\textit{Corresponding Author}}}

\begin{itemize}
    \item DataFlow Team: Project Team
    \item Bohan Zeng: \CoreContributor; Defines the scope of world model tasks and designs the OpenWorldLib framework.
    \item Daili Hua: \CoreContributor; Develops the pipeline for interactive video/audio generation and reasoning tasks in OpenWorldLib.
    \item Kaixin Zhu: \CoreContributor; Develops the pipeline for 3D generation tasks in OpenWorldLib.
    \item Yifan Dai: \CoreContributor; Develops the pipeline for VLA and reasoning tasks in OpenWorldLib.
    \item Bozhou Li: \CoreContributor; Develops the pipeline for VLA and reasoning tasks in OpenWorldLib.
    \item Yuran Wang: \CoreContributor; Develops the pipeline for interactive video generation tasks in OpenWorldLib.
    \item Chengzhuo Tong: \CoreContributor; Develops the pipeline for interactive video generation tasks in OpenWorldLib.
    \item Yifan Yang: \CoreContributor; Develops the pipeline for simulator and 3D generation tasks in OpenWorldLib.
    \item Mingkun Chang: \CoreContributor; Develops the pipeline for interactive video generation tasks in OpenWorldLib.
    \item Jianbin Zhao: \CoreContributor; Develops the pipeline for 3D generation tasks in OpenWorldLib.
    \item Zhou Liu: \Contributor; Designs world model task demonstrations and promotes the framework.
    \item Hao Liang: \Contributor; Guides the design of the OpenWorldLib framework.
    \item Xiaochen Ma: \Contributor; Guides configuration optimization of the OpenWorldLib framework.
    \item Ruichuan An: \Contributor; Provides knowledge assistance for world model-related tasks.
    \item Junbo Niu: \Contributor; Provides technical guidance on world model reasoning and memory.
    \item Zimo Meng: \Contributor; Tests the OpenWorldLib pipelines.
    \item Tianyi Bai: \Contributor; Provides technical guidance on world model reasoning.
    \item Meiyi Qiang: \Contributor; Promotes the framework and tests pipelines for OpenWorldLib.
    \item Huanyao Zhang: \Contributor; Promotes the framework and tests pipelines for OpenWorldLib.
    \item Zhiyou Xiao: \Contributor; Tests the OpenWorldLib pipelines.
    \item Tianyu Guo: \Contributor; Develops the pipeline for audio generation and understanding tasks in OpenWorldLib.
    \item Qinhan Yu: \Contributor; Tests the OpenWorldLib pipelines.
    \item Runhao Zhao: \Contributor; Tests the OpenWorldLib pipelines.
    \item Zhengpin Li: \Contributor; Tests the OpenWorldLib pipelines.
    \item Xinyi Huang: \Contributor; Develops the pipeline for interactive video generation tasks in OpenWorldLib.
    \item Yisheng Pan: \Contributor; Tests the OpenWorldLib pipelines.
    \item Yiwen Tang: \Contributor; Provides technical support for world model-related tasks.
    \item Juanxi Tian: \Contributor; Provides technical guidance on unified reasoning and generation.
    \item Yang Shi: \Contributor; Provides technical support for the design of the framework.
    \item Yue Ding: \Contributor; Provides technical support for world model reasoning.
    \item Xinlong Chen: \Contributor; Provides technical support for world model reasoning.
    \item Hongcheng Gao: \Contributor; Provides technical support for world model simulator.
    \item Minglei Shi: \Contributor; Promotes the framework and tests pipelines for OpenWorldLib.
    \item Jialong Wu: \Contributor; Provides knowledge guidance for world model-related tasks.
    \item Zekun Wang: \Contributor; Promotes the framework and tests pipelines for OpenWorldLib.
    \item Yuanxing Zhang: \ProjectSupervisor; Guides the design of the OpenWorldLib framework and supervises the project.
    \item Xintao Wang: \Contributor; Provides technical guidance on world model-related tasks.
    \item Pengfei Wan: \Contributor; Provides technical guidance on world model synthesis.
    \item Yiren Song: \Contributor; Provides technical guidance on world model synthesis and VLA.
    \item Mike Zheng Shou: \Contributor; Provides technical guidance on world model synthesis and VLA.
    \item Wentao Zhang: \CorrespondingAuthor, \ProjectSupervisor; Writes the manuscript and supervises the project.
\end{itemize}

\end{document}